\documentclass[preprint,12pt]{elsarticle}

\usepackage{amssymb}
\usepackage{amsmath}
\usepackage{booktabs}
\usepackage{multirow}
\usepackage{subcaption}
\usepackage{hyperref}

\journal{Information Fusion}

\begin{document}

\begin{frontmatter}



\title{Psychologically-Grounded Graph Modeling \\for Interpretable Depression Detection}


\author[inst1]{Rishitej Reddy Vyalla\fnref{equal}}

\author[inst2]{Kritarth Prasad\fnref{equal}}

\author[inst2]{Avinash Anand\corref{cor1}}
\ead{Avinash.Anand@singaporetech.edu.sg}

\author[inst3]{Erik Cambria}

\author[inst4,inst5]{Shaoxiong Ji}

\author[inst6]{Faten S. Alamri}

\author[inst2]{Zhengkui Wang}

\affiliation[inst1]{organization={IIIT Delhi},
            city={New Delhi},
            country={India}}

\affiliation[inst2]{organization={Singapore Institute of Technology},
            country={Singapore}}

\affiliation[inst3]{organization={Nanyang Technological University},
            country={Singapore}}

\affiliation[inst4]{organization={ELLIS Institute Finland},
            city={Espoo},
            country={Finland}}
            
\affiliation[inst5]{organization={University of Turku},
            city={Turku},
            country={Finland}}

\affiliation[inst6]{organization={Princess Nourah bint Abdulrahman University},
            city={Riyadh},
            country={Saudi Arabia}}

\cortext[cor1]{Corresponding author}
\fntext[equal]{These authors contributed equally to this work.}





\begin{abstract}
Automatic depression detection from conversational interactions holds significant promise for scalable screening but remains hindered by severe data scarcity and a lack of clinical interpretability. Existing approaches typically rely on black-box deep learning architectures that struggle to model the subtle, temporal evolution of depressive symptoms or account for participant-specific heterogeneity. In this work, we propose PsyGAT (Psychological Graph Attention Network), a psychologically grounded framework that models conversational sessions as dynamic temporal graphs. We introduce \emph{Psychological Expression Units (PEUs)} to explicitly encode utterance-level clinical evidence, structuring the session graph to capture transitions in psychological states rather than mere semantic dependencies. To address the critical class imbalance in depression datasets, we employ clinically approved persona-based data augmentation, enable robust model learning. Additionally, we integrate session-level personality context directly into the graph structure to disentangle trait-based behavior from acute depressive symptoms. PsyGAT achieves state-of-the-art performance, surpassing both strong graph-based baselines and closed-source LLMs like GPT-5, achieving 89.99 and 71.37 Macro F1 scores in DAIC-WoZ and E-DAIC, respectively. We further introduce Causal-PsyGAT, an interpretability module that identifies symptom triggers. Experiments show a 20\% improvement in MRR for identifying causal indicators, effectively bridging the gap between depression monitoring and clinical explainability. The full augmented dataset is publicly available at \url{https://doi.org/10.6084/m9.figshare.31801921}.
\end{abstract}




\end{frontmatter}



\section{Introduction}
Major Depressive Disorder (MDD) is a prevalent mental health condition that disrupts cognitive, emotional, and social functioning. Early identification of depressive symptoms is critical for timely intervention, yet clinical screening remains resource-intensive and difficult to scale. These constraints have motivated computational approaches that leverage signals from conversational interactions, including language~\cite{teferra2025llmdepression}, acoustic speech characteristics~\cite{maji2025speechfm}, and broader behavioral indicators~\cite{liu2025emoscan}, to support scalable and data-driven screening pipelines.



\begin{figure*}[!t]
  \centering
  \includegraphics[width=\textwidth]{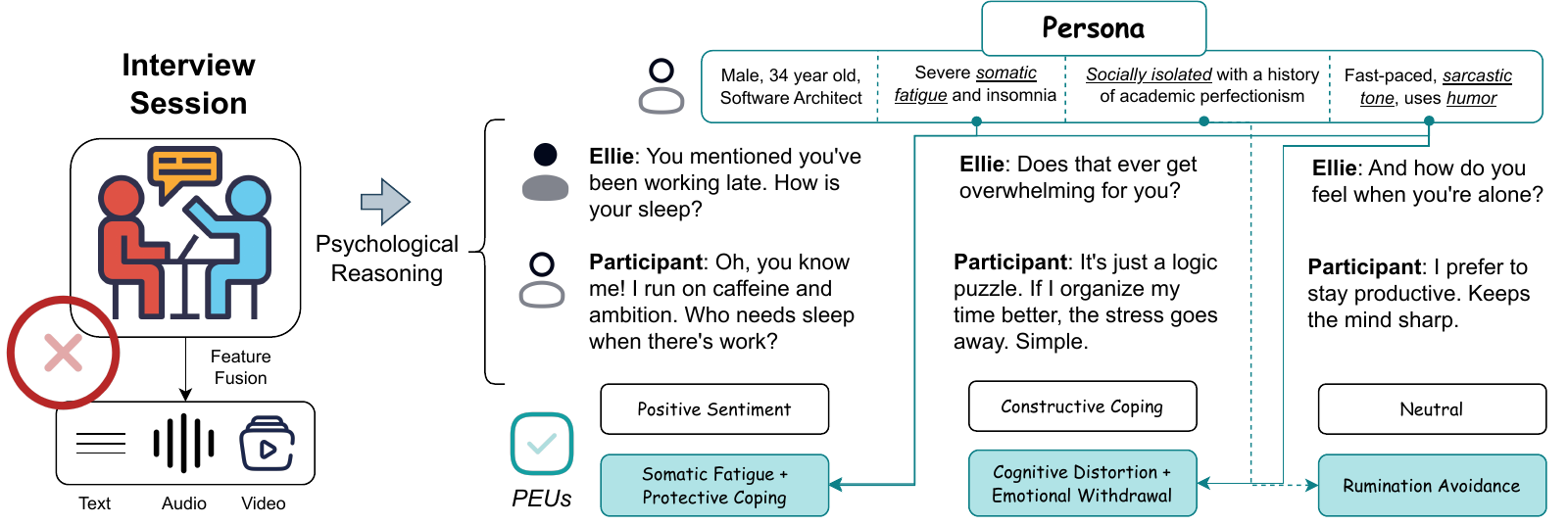}
  \caption{\textbf{Feature Fusion vs. Psychological Reasoning} (left) relies on surface semantics, failing to detect depression in high-functioning individuals who mask their symptoms. In contrast, our approach (right) integrates Persona profiles to interpret text through a clinical lens, extracting Psychological Expression Units (PEUs) that reveal hidden symptoms like Somatic Fatigue and Protective Coping for accurate depression prediction.}
  \label{fig:psy_reasoning}
\end{figure*}

Despite steady progress in depression assessment, three critical bottlenecks continue to limit the reliability of existing methods. \textbf{1)} Depressive signals in conversational settings are inherently subtle and transient. Unlike consistent emotional polarity, clinical symptoms ~\cite{shreevastava-foltz-2021-detecting} ranging from specific cognitive distortions and expressions of hopelessness to fleeting mentions of somatic fatigue or social withdrawal, manifest sporadically within a dialogue. These critical symptomatic moments often occur irregularly amidst largely neutral conversation, meaning standard models that aggregate features over long durations risk diluting their diagnostic value. \textbf{2)} Current baselines largely overlook the influence of participant persona ~\cite{cognitionwdepression, fu2025mpddchallengemultimodalpersonalityaware}. Depression is a highly heterogeneous disorder, and its behavioral manifestation is deeply intertwined with an individual's baseline personality, as illustrated in Figure \ref{fig:psy_reasoning}, we show how depressive signals manifest differently depending on an individual's personality traits. By failing to condition predictions on unique persona traits, standard approaches struggle to disentangle stable personality characteristics from acute depressive symptoms. \textbf{3)} These modeling challenges are exacerbated by severe data scarcity. The field relies heavily on datasets like DAIC-WOZ~\cite{2404.14463v1} \& E-DAIC, which contain only approximately $\sim$260 training samples and $\le$57 depressed participants. This paucity and class imbalance fundamentally constrain the complexity of learnable models, making them prone to overfitting and preventing the robust capture of the subtle, persona-dependent dynamics.

To address these challenges, we propose a psychologically grounded graph-based framework based on Graph Attention Network called PsyGAT (Psychological Graph Attention Network)  for session-level depression detection. PsyGAT consists of two main components: \emph{Psychological Expression Units} and Persona-aware Context Modeling, which capture the subtle and transient nature of depressive symptoms. First, we introduce PEUs, a set of discrete, clinically grounded symptomatic manifestations. Motivated by the Network Theory of Psychopathology \cite{networktheory, commentarynetwork}, which posits that mental disorders arise from direct causal interactions between specific symptoms rather than a latent disease entity, PEUs serve as the fundamental interacting nodes of our computational graph. The conceptualization of these nodes is further reinforced by Cognitive Appraisal Theory \cite{cogappraisal1, cogappraisal2}, which asserts that habitual negative evaluations, such as the persistent underestimation of one's coping potential, serve as critical cognitive antecedents that precipitate stable depressive states. Recent advancements in natural language processing validate this approach, demonstrating that incorporating cognitive appraisal dimensions provides a robust computational paradigm for predicting complex emotional states directly from text \cite{wang2026appraisal}. To ensure high construct validity, these units are extracted using a strict low-inference behavioral observation protocol; this ensures we capture concrete, explicitly stated cognitive distortions aligned with Beck's cognitive theory \cite{beck1970cognitive, Allen2011beck, expcogbenk} without relying on subjective, high-inference interpretations that compromise data integrity. We model each session as a directed temporal graph where nodes fuse utterance semantics with PEU signals, while edges explicitly encode the causal evolution of psychological states across adjacent utterances. This structure allows the model to isolate and track symptom dynamics that are otherwise lost in flat sequential architectures. Second, we employ a Persona-Aware Context Modeling mechanism to refine the temporal evolution of these signals. Grounded in the Diathesis-Stress Model \cite{Broerman2017} which conceptualizes baseline personality as a predisposing vulnerability (diathesis) that moderates the onset of depressive episodes \cite{Klein2011PersonalityAD} and research on affect dynamics, which demonstrates that these traits critically modulate an individual's emotional inertia and vulnerability to symptom transitions, our approach moves beyond treating personality as a static feature. Instead, we use participant-specific persona traits (see Figure \ref{fig:psy_reasoning}) to explicitly modulate the edge weights governing PEU transitions. This allows the model to learn how a specific personality type influences the progression of psychological states across adjacent utterances, effectively disentangling trait-based behavioral patterns from acute depressive dynamics. Furthermore, to mitigate the critical bottlenecks of data scarcity and class imbalance, we implement a persona-driven LLM-based data augmentation framework with few-shot anchoring, and we synthesize high-fidelity clinical sessions conditioned on diverse persona profiles. This ensures that the training data captures a wide spectrum of symptom expressions while preserving the structural integrity of clinical interviews. Finally, we bridge the gap between prediction and transparency through Causal-PsyGAT, an interpretability module that constructs a causal graph over the learned PEU transitions. Unlike standard attention maps which only show correlations, this framework identifies directed cause-effect relationships, enabling us to rank the specific causal triggers that precipitate distinct PEU symptoms. This granularity allows us to rigorously trace the trajectory of depressive signals, identifying both the retrospective origins of observed behaviors and their prospective contribution to the final diagnostic decision.

\paragraph{The main contributions of this paper are as follows:}
\begin{itemize}
    \item This study proposes PsyGAT, a novel graph attention framework that integrates \emph{Psychological Expression Units (PEUs)} with Persona-Aware Context Modeling. By explicitly encoding clinical evidence into nodes and modulating edge transitions based on personality traits, PsyGAT effectively disentangles trait-based behaviors from acute depressive symptoms, capturing the dynamic evolution of psychological states.
    \item To mitigate the severe scarcity and class imbalance of datasets like DAIC-WOZ, we introduce a persona-driven LLM augmentation strategy. By synthesizing sessions grounded in diverse clinical profiles, we enhance data diversity and training robustness without compromising conversational structure.
    \item Experiment results demonstrate that PsyGAT significantly outperforms strong graph-based baselines and closed-source powerful LLMs like GPT-5 on the DAIC-WOZ and E-DAIC benchmarks, achieving Macro F1 scores of 89.99 and 71.37, respectively.
    \item Finally, Causal-PsyGAT, an interpretability module that constructs causal graphs from learned PEU transitions. Unlike standard attention maps, this framework identifies directed cause-effect relationships for symptom triggers, achieving a 20\% improvement in MRR and offering a transparent pathway for clinical explainability.
\end{itemize}

\section{Related Work}

\subsection{Datasets for Depression Detection}

The DAIC-WOZ dataset is the most commonly used benchmark for automatic depression detection from clinical interviews, providing multimodal recordings and session-level depression labels~\cite{s44184-024-00112-8}. Owing to the lack of large-scale alternatives, it has been widely adopted despite known limitations such as small sample size and pronounced class imbalance.

Recent analyses have identified dataset-specific biases in DAIC-WOZ, particularly the influence of interviewer prompts, which can enable shortcut learning and inflate performance without capturing participant-centered depressive signals~\cite{2404.14463v1}. Related critiques further argue that many reported results on DAIC-WOZ may not generalize beyond dataset-specific artifacts, underscoring the need for more robust evaluation protocols and representations grounded in participant behavior rather than interviewer structure.

To address these limitations, several recent efforts have introduced new clinically grounded datasets. For example, the PDCH dataset provides real-world depression consultations with aligned speech, text, and HAMD-17 assessments, offering a higher-fidelity alternative for training and evaluation~\cite{cao2025pdch}. Other multimodal datasets, such as MODMA, incorporate physiological and acoustic signals to broaden the scope of depression-related behavioral evidence. These developments motivate approaches that reduce reliance on superficial cues while addressing data scarcity through structured modeling and augmentation.

\subsection{Depression Detection Methods}

Prior work on depression detection spans unimodal and multimodal approaches using text, audio, and visual cues~\cite{computation-13-00009,Li2025PredictingDB}. Multimodal methods generally outperform unimodal ones by combining complementary linguistic and paralinguistic information, often through early or late fusion strategies.

Recent deep learning models employ transformers, graph neural networks, and foundation models to integrate multimodal signals and capture long-range dependencies in interviews~\cite{s11042-023-18079-7,gomezzaragoza2025foundation,maji2025speechfm}. Multi-instance learning formulations have also been explored to address session-level labeling and temporal sparsity, demonstrating improved robustness under limited supervision~\cite{zhang2025mildepression}.

Despite strong empirical performance, many existing approaches rely on high-dimensional latent representations with limited interpretability and only implicit modeling of conversational structure. Moreover, temporal dynamics are often captured through generic sequence models rather than representations aligned with psychological theory. These limitations highlight the need for methods that explicitly encode conversational structure and model interactions between psychologically meaningful dimensions over time.

\subsection{Graph-based Modeling for Depression Analysis}

Beyond sequential encoders, recent work has explored \emph{graph} representations to better reflect the non-linear structure of clinical interviews and to enable more interpretable transcript summaries. Agarwal et al.~\cite{agarwal2024multiviewgraph} construct transcript graphs that connect utterances via semantic similarity and also build keyword-correlation graphs that inject corpus-level topical structure; importantly, they introduce a \emph{multi-view} graph formulation that separates interviewer questions and participant answers while explicitly modeling cross-view links, yielding strong performance on DAIC-WOZ-style binary depression classification. In parallel, graph learning has also been applied to multimodal depression-related prediction in neuroimaging, where GNNs over fMRI/EEG connectivity graphs are fused via cross-modality correlation objectives and paired with interpretability analyses to identify salient subnetworks for outcome prediction~\cite{jiao2025deepgraph}. Collectively, these graph baselines motivate structure-aware modeling that can encode relational dependencies (within and across conversational roles or modalities) rather than relying solely on generic sequence dynamics.

\subsection{Personality and Context-aware Modeling}

Recent work has highlighted the importance of personality and persona information for contextualizing affective and mental health-related language understanding. In recent work, Personalized sentiment analysis framework has been proposed that models user characteristics across multiple levels, including personality traits, demonstrating that personality-aware conditioning can improve sentiment interpretation by accounting for individual differences in perception~\cite{personalized-sentiment-analysis}. This line of work emphasizes that identical linguistic expressions may carry different affective meanings depending on stable personal attributes.

In mental health-oriented dialogue systems, persona-aware modeling has been shown to improve both user engagement and interpretability. SupportPlay introduces persona-aware emotional support agents that leverage user profiles and long-term memory to deliver personalized responses~\cite{supportplay}. Similarly, recent LLM-based mental health systems condition generation and assessment on user traits and historical context, illustrating the value of structured persona representations beyond surface-level conversation history~\cite{liu2025emoscan}.

These findings motivate the inclusion of personality as a session-level contextual factor, complementing utterance-level behavioral signals and enabling more individualized modeling of mental health states across conversational trajectories.

\subsection{Explainable Depression Detection}

Explainability has received increasing attention in depression detection due to the high-stakes nature of mental health decision-making. Early efforts incorporate interpretable auxiliary signals and model their interactions with textual representations using relational reasoning frameworks. 


More recent work focuses on designing intrinsically explainable neural architectures. Recent studies introduced hierarchical attention network for depression detection on social media, where attention mechanisms operate at the context level to identify salient posts and psychologically meaningful features such as metaphor concept mappings~\cite{1_1}. Beyond attention-based explanations, causal and fairness-aware frameworks have been proposed to disentangle sensitive attributes and provide structured explanations for multimodal depression predictions~\cite{cheong2024fairrefuse}.

Despite these advances, most explainable methods rely on attention weights or predefined linguistic features and often operate at the post or user level without explicitly modeling temporal psychological dynamics. This motivates structured approaches that support causal, relational, and transition-based explanations aligned with evolving mental states over the course of an interview.

\begin{figure*}[!t]
  \centering  \includegraphics[width=\textwidth]{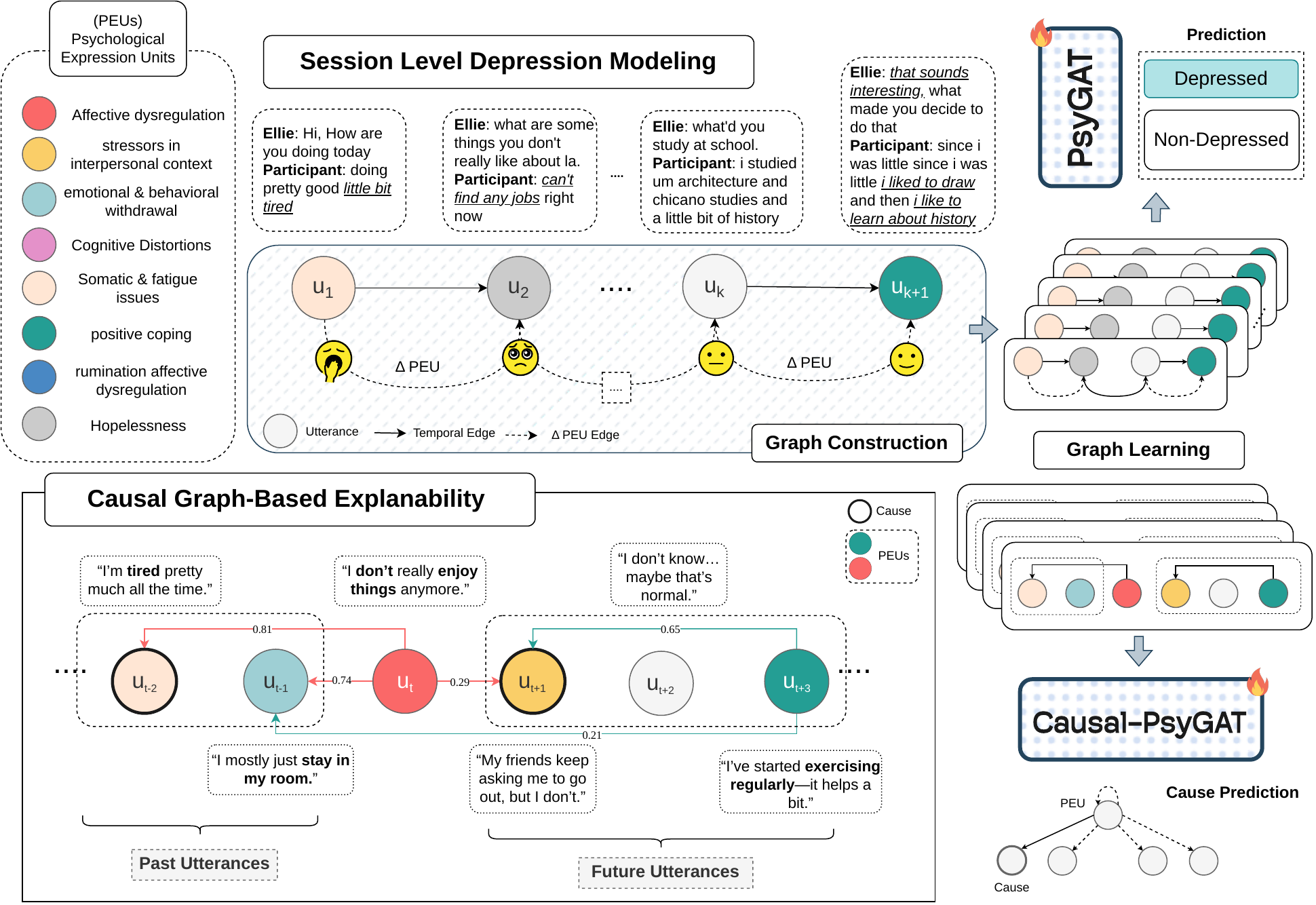}
  \caption{\textbf{End-to-end framework for explainable depression detection}}
  \label{fig:basic_pipeline}
\end{figure*}


\section{Methodology}






The overall pipeline (Figure \ref{fig:basic_pipeline}) consists of four stages:
\begin{enumerate}
    \item \textbf{Dataset Construction and Augmentation} : Combining real benchmarks with persona-conditioned synthetic data to address scarcity and class imbalance.
    \item \textbf{Psychological Expression Unit (PEU)}: construction as psychologically grounded utterance-level signals.
    \item \textbf{Session Graph Representation}: Directed temporal graph representation integrating semantic and psychological features.
    \item \textbf{Persona-Aware Context Modeling}: Contextual extension of conversational sessions with persona-aware modelling.
    \item \textbf{Causal Graph-Based Explainability}: Attributing detected PEUs to their conversational antecedents via causal edge classification over the session graph.
\end{enumerate}




\subsection{Dataset Construction and Augmentation}

We construct our training, validation, and test sets using a combination of original DAIC-WOZ sessions and augmented conversational sessions. All sessions are processed into a unified representation to ensure consistency across original and augmented data. Augmentation is performed under strict structural and domain constraints to preserve the conversational format and psychological plausibility of the original interviews.

\subsubsection{DAIC-WOZ Sessions}
We use conversational sessions from the DAIC-WOZ dataset, which consists of semi-structured clinical interviews between a virtual interviewer and human participants. Each session comprises a sequence of utterances from both the participant and the interviewer and is annotated with a session-level depression label derived from standardized clinical assessments. DAIC-WOZ serves as the primary source of real-world data in our experiments.

\subsubsection{LLM-based Data Augmentation}

To alleviate the limited size and class imbalance in DAIC-WoZ and E-DAIC clinical interview dataset, we augment the training data using a large language model (LLM). Instead of paraphrasing existing utterances \cite{chen-etal-2024-depression}, we generate additional synthetic sessions conditioned on clinically grounded depression-oriented personas adapted from TalkDep \cite{talkdep}. These personas specify demographic attributes and characteristic symptom patterns, providing structured guidance for generation.
To ensure principled augmentation, we impose strict generation rules that are subsequently verified through manual human inspection of a representative sample. Specifically, we validated that (1) The dialogue remains original and structurally aligned with the source data. (2) The interaction maintains a safe, therapeutic tone without medical inconsistencies and (3) participant responses strictly adhere to the assigned persona. Crucially, human validators explicitly assess intra-class diversity, ensuring varied linguistic and behavioral manifestations within both depressed and non-depressed classes to prevent the model from learning a singular prototypical profile. Ultimately, guided by these rigorous constraints, less than 1\% of the synthetic sessions failed human evaluation and required regeneration.

\subsection{Psychological Expression Units (PEUs)}

We introduce \emph{Psychological Expression Units (PEUs)} as a structured representation of psychologically salient evidence expressed in conversational utterances. PEUs are designed to capture \emph{explicit} indicators of depressive cognition, affect, behavior, and coping in a way that is both operationalizable and compatible with graph-based modeling. Importantly, PEUs do not represent inferred mental states, but only explicit, text-supported psychological evidence expressed at the utterance level.

\subsubsection{PEU Taxonomy and Definitions}

Each utterance is analyzed for the presence of evidence belonging to one or more PEU categories. We use eight categories:
(i) cognitive distortions,
(ii) hopelessness/helplessness,
(iii) self-negativity,
(iv) stressors and interpersonal context,
(v) emotional and behavioral withdrawal,
(vi) somatic fatigue and sleep issues,
(vii) rumination and affective dysregulation, and
(viii) protective or positive coping.

For each utterance, PEUs are extracted only when the utterance text provides explicit support, and the extracted evidence is represented as short phrases (verbatim spans) associated with the corresponding category. This yields a fine-grained mapping from utterances to psychologically meaningful evidence without aggregating across utterances, enabling downstream models to capture \emph{temporal evolution} of psychological signals over the session.

\subsubsection{PEU Tensor Representation}

To use PEUs as model inputs, we convert utterance-level PEU annotations into fixed-order tensors aligned with the session timeline. 

For a session with $T$ utterances and $C=8$ PEU categories,
we construct a PEU tensor
\[
\mathbf{P} = [\mathbf{p}_1, \dots, \mathbf{p}_T]^\top \in \mathbb{R}^{T \times C},
\]
where $\mathbf{p}_t \in \mathbb{R}^{C}$ corresponds to the PEU representation of utterance $u_t$.

Each time step corresponds to one utterance, and each dimension corresponds to a PEU category in a fixed canonical order.

We represent PEUs as a fixed-dimensional utterance-level vector, where each dimension corresponds to a predefined psychological expression category. For each utterance, a dimension is set to $1$ if a phrase associated with that category is present, and $0$ otherwise. 

The protective or positive coping dimension takes values in $\{-1, 0, 1\}$, indicating mitigating/protective coping expression, absence, or positive coping expression, respectively.

These PEU tensor representations provide a structured and temporally aligned interface between psychologically grounded evidence extraction and the graph-based learning framework described in subsequent sections.


\subsection{Session Graph Representation}

As shown in Figure \ref{fig:basic_pipeline}, each conversational session is represented as a directed graph:
\[
G = (V, E),
\]
where each node $v_t \in V$ corresponds to utterance $u_t$ in the session.

\subsubsection{Node Features and Edge Construction}

Each utterance $u_t$ is represented as a node $v_t$ with feature vector
\[
\mathbf{x}_t = [\mathbf{s}_t \,\Vert\, \mathbf{p}_t],
\]
where $\mathbf{s}_t$ is a dense semantic embedding obtained from a pre-trained sentence encoder, and $\mathbf{p}_t$ is the corresponding PEU vector. This joint representation integrates linguistic content with psychologically grounded evidence at the node level.

To capture conversational dynamics, we introduce directed edges between temporally adjacent utterances $(v_t, v_{t+1})$. Each edge is associated with an attribute vector
\[
\mathbf{e}_{t,t+1} = \mathbf{p}_{t+1} - \mathbf{p}_t,
\]
which encodes changes in psychological expression across consecutive turns. By explicitly modeling PEU differences as edge attributes, the graph captures transitions in psychological state directly, rather than relying solely on node-level aggregation.

\subsection{Persona-Aware Context Modeling}

Depressive symptom expression varies substantially across individuals due to demographic background, symptom severity, and communication style. To capture such inter-individual variability, we incorporate explicit persona-level information into the graph modeling pipeline.

Following the persona construction principles of TalkDep~\cite{talkdep}, we define four predefined personas representing distinct demographic profiles, symptom severity patterns, and conversational behaviors. Each persona specification encodes (i) basic personal attributes, (ii) characteristic negative symptoms and their relative prominence, (iii) contextual life background, and (iv) stylistic tendencies in emotional expression. 

For each session, a persona label $p \in \{1,\dots,K\}$ (with $K=4$) is assigned and treated as a session-level prior. Let $G = (V,E)$ denote the session graph constructed from utterances. After graph attention propagation, we obtain contextualized node representations $\mathbf{h}_t$ and compute a session-level embedding
\[
\mathbf{h}_G = \mathrm{READOUT}(\{\mathbf{h}_t\}_{t=1}^{T}),
\]
where $\mathrm{READOUT}(\cdot)$ denotes mean or attention-based pooling.

We then map the persona label to a learnable embedding vector
\[
\mathbf{z}_p = \mathrm{Embed}(p),
\]
and incorporate persona information by conditioning the session representation:
\[
\tilde{\mathbf{h}}_G = [\mathbf{h}_G \,\Vert\, \mathbf{z}_p].
\]

The final prediction is computed as
\[
\hat{y} = \mathrm{MLP}(\tilde{\mathbf{h}}_G).
\]

Compared to persona-agnostic modeling ($\hat{y}=\mathrm{MLP}(\mathbf{h}_G)$), this formulation introduces a persona-dependent prior that shifts the decision boundary according to individual-level characteristics. While the graph structure and message passing remain unchanged, the downstream prediction becomes explicitly conditioned on persona information, enabling the model to account for systematic differences in symptom manifestation.

\subsection{Causal Graph-Based Explainability}

We implement this component within our proposed framework, \textbf{Causal-PsyGAT}, illustrated in Figure~\ref{fig:basic_pipeline}.

While session-level predictions indicate whether a participant is depressed,
clinical interpretability requires understanding \emph{how} psychological symptoms emerge over the course of a conversation.

We therefore frame explainability as a causal attribution problem: given an utterance expressing a psychological symptom, identify preceding utterances that plausibly contributed to its emergence.

\subsubsection{Problem Motivation}

From a psychological perspective, appraisal theory posits that emotional and cognitive states arise from individuals’ evaluations of events and evolve as these appraisals change over time \cite{roseman2001appraisal,scherer1999appraisal}. In conversational settings, utterances can be viewed as discrete events that trigger or modify underlying appraisals, leading to the gradual emergence of observable psychological symptoms. Understanding depression, therefore, requires not only identifying \emph{what} symptoms are present, but also \emph{understanding which preceding conversational events} contributed to their development.

However, most existing explainability approaches for depression detection emphasize correlational signals, such as salient words, attention weights, or post-hoc feature attributions. While these methods can highlight associations, they do not explicitly account for the temporal evolution of psychological symptoms or the event-driven nature of conversational context emphasized by appraisal-based theories.

Our approach addresses this gap by framing explainability as a causal attribution problem over psychological symptoms. Instead of explaining predictions via raw text features or internal attention mechanisms, we identify prior conversational events that contribute to the emergence or modulation of specific psychological expressions, thereby modeling symptom development as a temporally grounded causal process.

\subsubsection{Causal Attribution Framework}
Given a conversational session represented as a sequence of utterances, and a target utterance in which one or more PEUs are expressed, the goal of causal attribution is to identify a subset of preceding utterances that are most likely to have influenced the emergence of the target PEUs.

Let a conversational session be represented as an ordered sequence of participant utterances
\[
\mathcal{U} = \{u_1, u_2, \dots, u_T\},
\]
where each utterance $u_t$ is associated with a learned representation
\[
\mathbf{h}_t \in \mathbb{R}^{d},
\]
obtained from the trained session-level graph model described in the previous section.

Each utterance may express one or more Psychological Expression Units (PEUs). For a given PEU instance occurring at utterance index $t$, the goal of causal attribution is to identify which surrounding utterances plausibly contributed to the emergence of that PEU.

Candidate Cause Selection. For a target PEU instance at utterance $u_t$ (where $u_t$ denotes the $t$-th utterance in the session), we define a local temporal context window
\[
C_t = \{u_j \mid j \in [t-w, t+w],\ j \neq t\},
\]
where $u_j$ denotes the utterance at position $j$ in the same session and $w$ is a fixed window size. In our experiments, we set $w = 5$. Each utterance $u_j \in C_t$ is treated as a candidate causal antecedent.

Causal Graph Construction. For each PEU instance, we construct a directed graph
\[
G_t = (V_t, E_t),
\]
where the node set
\[
V_t = \{v_0, v_1, \ldots, v_{N_t}\}
\]
consists of: (i) a target node $v_0$ representing the PEU instance at utterance $u_t$, and (ii) candidate nodes $v_i$ corresponding to utterances $u_{j_i} \in C_t$.
The edge set is given by
\[
E_t = \{(v_0, v_i) \mid i = 1, \ldots, N_t\},
\]
corresponding to directed edges from the PEU node to each candidate utterance node.

Each edge $(v_0, v_i)$ is associated with a binary label indicating whether utterance $u_{j_i}$ is annotated as a causal contributor to the PEU at $u_t$.

\paragraph{Training Objective}
The model is trained using supervised edge-level classification. To address class imbalance between causal and non-causal candidates, we employ a weighted focal loss:
\[
\mathcal{L} =
\frac{1}{|E|} \sum_{t,i}
\alpha (1-\hat{y}_{t,i})^{\gamma}
\,\mathrm{BCE}(\hat{y}_{t,i}, y_{t,i}),
\]
where $\alpha$ and $\gamma$ are focal loss hyperparameters and $\mathrm{BCE}$ denotes binary cross-entropy.

\paragraph{Explanation and Evaluation}
At inference time, candidate utterances are ranked by their predicted causal probabilities $\hat{y}_{t,i}$. Explanation quality is evaluated using ranking-based metrics, including Hit@K and Mean Reciprocal Rank (MRR), which assess whether true causal utterances appear among the top-ranked candidates.

This causal attribution model is trained only after the session-level depression prediction model has converged. It operates in a post hoc manner and does not influence prediction performance directly, but instead provides structured explanations of how psychological expressions emerge over the course of a conversation.

\begin{table*}[t]
\centering
\small
\setlength{\tabcolsep}{6pt}
\renewcommand{\arraystretch}{1.2}
\makebox[\textwidth][c]{%
\begin{tabular}{cccccccccc}
\toprule
\multirow{2}{*}{Method} &
\multicolumn{3}{c}{Element} &
\multicolumn{3}{c}{DAIC-WOZ} &
\multicolumn{3}{c}{E-DAIC} \\
\cmidrule(lr){2-4} \cmidrule(lr){5-7} \cmidrule(lr){8-10}
 & T & Q & P
 & Depressed & Control & Macro
 & Depressed & Control & Macro \\
\midrule
\multicolumn{10}{c}{\textbf{LLM Baselines}} \\
\midrule
Gemma3-4B &
$\circ$ & $\circ$ & $\times$ &
53.33 & 0.00 & 26.67 &
36.92 & 8.89 & 22.91 \\

Qwen2.5-Omni-7B &
$\circ$ & $\circ$ & $\times$ &
66.67 & 81.97 & 74.32 &
38.46 & 80.56 & 59.71 \\

Qwen3-Omni-30B &
$\circ$ & $\circ$ & $\times$ &
65.00 & 74.07 & 69.54 &
48.89 & 64.62 & 56.75 \\

Deepseek-v3.2 &
$\circ$ & $\circ$ & $\times$ &
68.75 & 70.59 & 69.67 &
52.16 & 65.62 & 58.90 \\

Kimi-k2-instruct &
$\circ$ & $\circ$ & $\times$ &
70.97 & 74.29 & 72.63 &
52.16 & 65.62 & 58.90 \\

GPT-5 &
$\circ$ & $\circ$ & $\times$ &
68.75 & 83.75 & 76.31 &
51.61 & 81.01 & 66.31 \\

\midrule
\multicolumn{10}{c}{\textbf{Our Approach}} \\
\midrule
\multirow{3}{*}{PsyGAT} &
$\circ$ & $\times$ & $\times$ &
83.33 & 90.48 & 86.90 &
45.54 & 86.36 & 65.91 \\

& $\circ$ & $\circ$ & $\times$ &
84.62 & 90.00 & 87.31 &
50.00 & 88.89 & 69.44 \\

& $\circ$ & $\circ$ & $\circ$ &
\textbf{84.96} & \textbf{93.02} & \textbf{89.99} &
\textbf{52.63} & \textbf{90.11} & \textbf{71.37} \\

\bottomrule
\end{tabular}
    }
\caption{F1-score (\%) comparison on DAIC-WOZ and E-DAIC for LLM baselines and our method.
T, Q, and P denote the use of Patient Utterances, Therapist Questions, and Patient Persona, respectively.
Best results on DAIC-WOZ are highlighted in bold.}
\label{tab:llm_and_ours}
\end{table*}

\begin{table}[t]
\centering
\small
\setlength{\tabcolsep}{6pt}
\renewcommand{\arraystretch}{1.2}
\begin{tabular}{lccc}
\toprule
Method & Depressed & Control & Macro F1 \\
\midrule
$\omega$-GCN        & 78.26 & 89.36 & 83.81 \\
HCAG               & 76.92 & 86.36 & 81.64 \\
Similarity-MV      & --    & --    & 81.00 \\
KCG-MV             & --    & --    & 76.00 \\
\midrule
SEGA               & 81.48 & 88.37 & 84.93 \\
SEGA++             & 84.62 & 90.91 & 87.76 \\
\midrule
PsyGAT (Ours)      & \textbf{84.96} & \textbf{93.02} & \textbf{89.99} \\
\bottomrule
\end{tabular}
\caption{Comparison with graph-based baselines on the DAIC-WOZ dataset using Macro F1-score (\%). 
Results for Similarity-MV and KCG-MV are reported as Macro F1 only, as class-wise scores are not provided in the original work.}
\label{tab:graph_baselines}
\end{table}

\begin{table}[t]
\centering
\small
\setlength{\tabcolsep}{8pt}
\renewcommand{\arraystretch}{1.2}
\begin{tabular}{lcccc}
\toprule
Method & Hit@1 & Hit@3 & Hit@5 & MRR \\
\midrule
\textbf{LLM Baselines} \\
\midrule
GPT-5              &  32.46  &  58.43  &  62.91 &  45.68 \\
Qwen2.5-Omni-7B    &  20.00  &  37.43  &  46.91 &  30.11 \\
Qwen3-Omni-30B     &  30.22  &  47.01  &  52.83 &  39.69 \\

\midrule
\textbf{Graph-Based} \\
\midrule
Causal-PsyGAT (Ours) & \textbf{46.1} & \textbf{87.9} & \textbf{99.0} & \textbf{67.0} \\

\bottomrule
\end{tabular}
\caption{Explanation quality on the DAIC-WOZ dataset using ranking-based metrics.
Hit@K measures whether the ground-truth causal utterance appears among the top-K ranked candidates,
while MRR evaluates the mean reciprocal rank of the true causal utterance.}
\label{tab:causal_explanation_daic}
\end{table}

\section{Experimental Setup}

\subsection{Datasets}
We conduct our experiments on DAIC-WOZ and E-DAIC corpora for clinical depression detection. For DAIC-WOZ, we follow the official dataset partition. Because the test set labels are not publicly available, we report our evaluation metrics on the official development set. The original training partition comprises 107 interviews (30 depressed, 77 control), while the development set contains 35 interviews (12 depressed, 23 control). For the E-DAIC corpus, we adhere to the standard train-evaluation partition provided with the dataset. The original training set includes 163 interviews (37 depressed, 126 control), and the evaluation set consists of 56 interviews (12 depressed, 44 control).

To address data scarcity and the critical class imbalance inherent in both corpora, we employ a persona-based synthetic clinical approved interview strategy to expand our training pool. Specifically, we utilize the Kimi-k2-instruct \cite{kimiteam2026kimik2openagentic} Large Language Model (LLM) to generate synthetic training samples. For each augmented instance, the model is prompted with two key components: (i) a few-shot context comprising short conversational examples from the original dataset to preserve the semi-structured interviewer-participant format, and (ii) a target persona description drawn from 12 diverse TalkDep Personas, which explicitly constrains the simulated participant's communication style and depressive severity. Applying this protocol, we augment the DAIC-WOZ training set with 992 synthetic interview sessions (496 depressed, 496 control). Similarly, we generate 996 augmented samples for the E-DAIC training set (496 depressed, 500 control). We strictly isolate this augmented data, utilizing it exclusively during the training phase to prevent data leakage; all validation and testing are conducted solely on the original unmodified datasets.

\subsection{Experiment Setting}

We implement our models using PyTorch and DGL. Each patient utterance is encoded with a sentence transformer (\texttt{all-MiniLM-L6-v2}) to obtain a dense semantic vector. In our main experiment, we prepend the therapist's question to the patient's utterance before encoding, unless otherwise mentioned to explicitly remove it. Session graphs are constructed as directed temporal chains with edges between consecutive utterances. The edge attributes consist of the normalized PEU difference vectors between adjacent utterances.

For our graph encoder, we process the text and PEU embeddings through independent projection blocks, each consisting of Layer Normalization followed by a linear transformation to map them to a shared hidden dimension of 128. These separate streams are then summed and normalized to form the initial node features. To explicitly incorporate temporal edge attributes, we project the edge features and sum them into the corresponding destination nodes prior to each graph attention layer. Our core network relies on a two-layer GATv2 architecture with residual connections and a dropout rate of 0.20 with number of attention heads set to 2. Finally, a Set2Set readout (niters=4) aggregates the node embeddings into a session-level representation, which is passed through an MLP classification head to obtain the prediction logit.

We train graph models using AdamW with learning rate $2\times10^{-4}$ and weight decay $2\times10^{-4}$. Training is performed for up to 50 epochs with early stopping (patience $=8$) based on validation PR-AUC, and a ReduceLROnPlateau scheduler (factor $=0.5$, patience $=2$). Gradients are clipped to a maximum norm of 1.0. The main classification objective is focal loss ($\gamma=2.0$) by default, with binary cross-entropy as an alternative option. We additionally support a supervised contrastive auxiliary loss (InfoNCE; temperature $=0.2$), which is added to the main classification loss.

To reduce variance, we train an ensemble of five models with different random seeds and average predicted probabilities at inference. The final decision threshold is selected on the validation set via grid search to optimize either F1, F$_{0.5}$, or recall under a minimum precision constraint (0.75), depending on the reporting setting. We report depressed-class F1, control-class F1, and macro-F1, selecting the checkpoint based on validation PR-AUC.

\section{Results}

We evaluate our proposed framework on the DAIC-WOZ dataset and its extended variant (E-DAIC), and compare against strong large language model (LLM) baselines as well as representative graph-based approaches. Performance is reported using class-wise F1 scores and macro-F1, which are particularly appropriate under class imbalance.

\subsection{Depression Detection Performance}

Table~\ref{tab:llm_and_ours} reports the depression detection performance on DAIC-WOZ and E-DAIC. Among LLM baselines, GPT-5 achieves the strongest results on DAIC-WOZ with a macro-F1 of 76.31, followed closely by Qwen2.5-Omni-7B (74.32). However, all LLM baselines exhibit a substantial performance drop on E-DAIC, with macro-F1 scores ranging from 22.91 to 66.31. This consistent degradation indicates that prompt-based, zero-shot LLM approaches are highly sensitive to distributional shifts and dataset-specific characteristics, despite strong in-distribution performance.

In contrast, our graph-based PsyGAT framework achieves markedly stronger and more stable performance. Using utterance semantics alone already yields a macro-F1 of 86.90 on DAIC-WOZ. Incorporating therapist questions further improves performance to 87.31, while the full model with persona-aware context achieves the best overall results, reaching a macro-F1 of 89.99 on DAIC-WOZ and 71.37 on E-DAIC.

Notably, PsyGAT substantially outperforms all graph baselines on E-DAIC as shown in Table~\ref{tab:graph_baselines}, demonstrating improved cross-dataset generalization.
Overall, these results show that explicitly modeling psychologically grounded representations, therapist-patient interactions, and persona context enables more robust and generalizable depression detection than LLM-only approaches, particularly under distributional shift.

\subsection{Effect of Psychological and Contextual Components}

The ablation-style results in Table~\ref{tab:llm_and_ours} illustrate the incremental contribution of each component. Adding PEU-based representations improves discrimination for both depressed and control classes, while persona-aware conditioning yields the largest gains for the control class, indicating improved calibration and reduced false positives. This supports the hypothesis that stable session-level context can help disambiguate transient affective expressions from clinically relevant depressive patterns.

\subsection{Causal Explanation Quality}

Beyond predictive performance, we evaluate the quality of causal explanations produced by our framework. Table~\ref{tab:causal_explanation_daic} reports ranking-based metrics for identifying conversational utterances that causally contribute to specific psychological expression units in DAIC-WOZ.
We compare against LLM baselines because existing depression detection methods do not report utterance-level causal explanation metrics, nor evaluate explanation quality on DAIC-WOZ in a comparable, quantitative manner. Prior work typically relies on attention weights or post-hoc feature attribution, which does not provide explicit causal rankings and therefore cannot be evaluated using ranking-based metrics such as Hit@K or MRR.
PsyGAT substantially outperforms all LLM baselines across all metrics, achieving a Hit@1 of 55.61, Hit@3 of 94.26, and Hit@5 of 99.83. This indicates that the true causal utterance is correctly ranked within the top five candidates in nearly all cases, and within the top position in more than half of the instances. The high MRR of 74.09 further confirms that relevant causal antecedents are consistently ranked near the top. These results demonstrate that psychologically grounded graph modeling enables accurate, temporally coherent causal explanations beyond what is achievable with prompt-based LLM approaches.



\begin{figure}[t]
    \centering
    
    \begin{subfigure}[t]{0.48\linewidth}
        \centering
        \includegraphics[width=\linewidth]{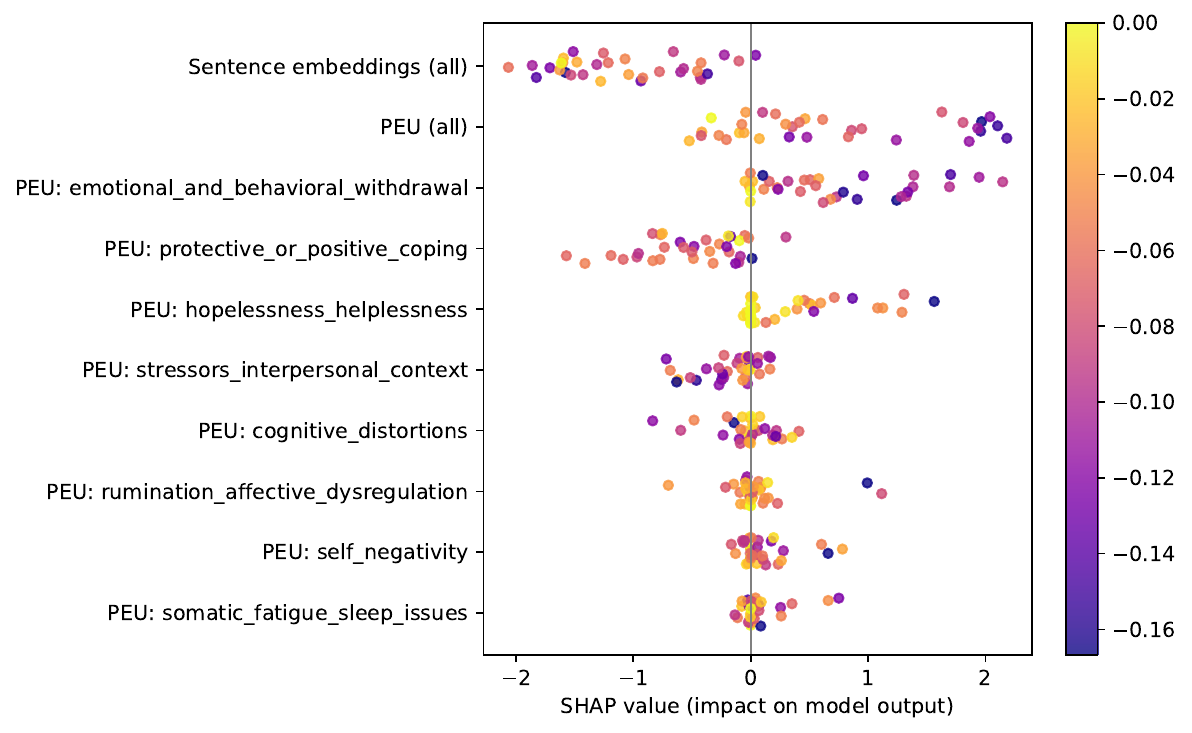}
        \caption{without persona}
        \label{fig:wo_persona}
    \end{subfigure}
    \hfill
    \begin{subfigure}[t]{0.48\linewidth}
        \centering
        \includegraphics[width=\linewidth]{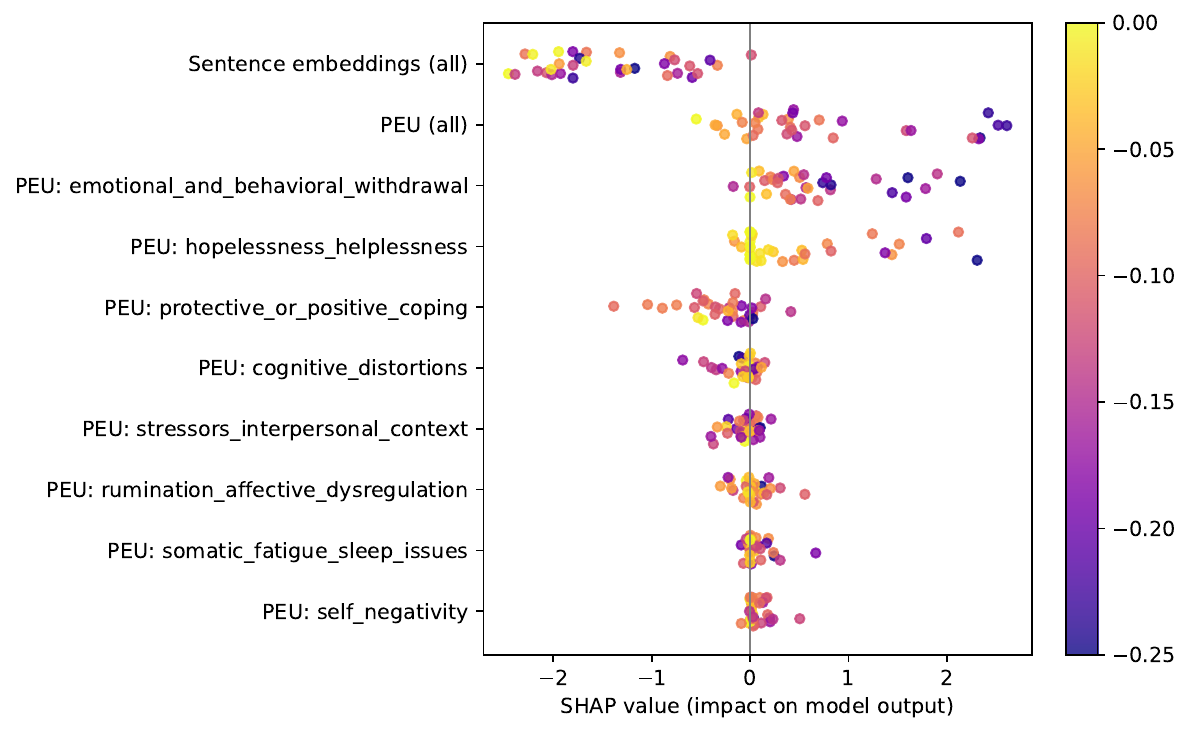}
        \caption{with persona}
        \label{fig:w_persona}
    \end{subfigure}
    
    \caption{SHAP analysis of individual feature group using without persona (a) and with persona (b) influence.}
    \label{fig:persona_comparison}
\end{figure}

\section{Ablation Study}

\subsection{Contribution of Individual Feature Groups}

To analyze the relative importance of different input modalities, we conduct an ablation study over individual feature groups and further examine their contributions using SHAP-based attribution. Specifically, we compare a baseline model without persona conditioning against the persona-aware variant.

Figure~\ref{fig:wo_persona} (without persona) and Figure~\ref{fig:w_persona} (with persona) illustrate the distribution of SHAP values across feature groups. Several systematic shifts are observed when persona conditioning is introduced.

First, the ranking of PEU categories changes. In the baseline model, \textit{protective\_or\_positive\_coping} is ranked above \textit{hopelessness\_helplessness}. After incorporating persona information, \textit{hopelessness\_helplessness} moves higher in importance, while \textit{protective\_or\_positive\_coping} drops in rank. This indicates that persona conditioning increases the model’s sensitivity to negative distress-related indicators relative to protective factors.

Second, sentence embeddings exhibit a stronger negative SHAP contribution in the persona-aware model. As shown in Figure~\ref{fig:w_persona}, the distribution of SHAP values for sentence embeddings shifts further toward the negative side compared to the baseline. This suggests that persona context amplifies how raw linguistic content contributes to high-risk predictions.

Third, low PEU values (blue points) for features such as \textit{PEU (all)} and \textit{emotional\_and\_behavioral\_withdrawal} are more tightly clustered on the positive SHAP side in the persona-aware model. This reflects a clearer separation between low-risk and high-risk cases, with low PEU activation more consistently pushing predictions toward the non-depressed class.

Finally, lower-ranked PEU categories (e.g., \textit{self\_negativity}) exhibit reduced variance around zero after persona integration. This suggests that persona conditioning concentrates explanatory power in the most diagnostically salient features, while diminishing reliance on weaker or noisier signals.

Overall, introducing persona information produces a polarizing effect on feature attribution: distress-related indicators are amplified, protective signals are relatively down-weighted, and minor features contribute less variance. This leads to sharper decision boundaries and improved interpretability of the model’s risk assessment behavior.

\begin{figure*}[ht]
\begin{center}
\centerline{\includegraphics[width=1\linewidth]{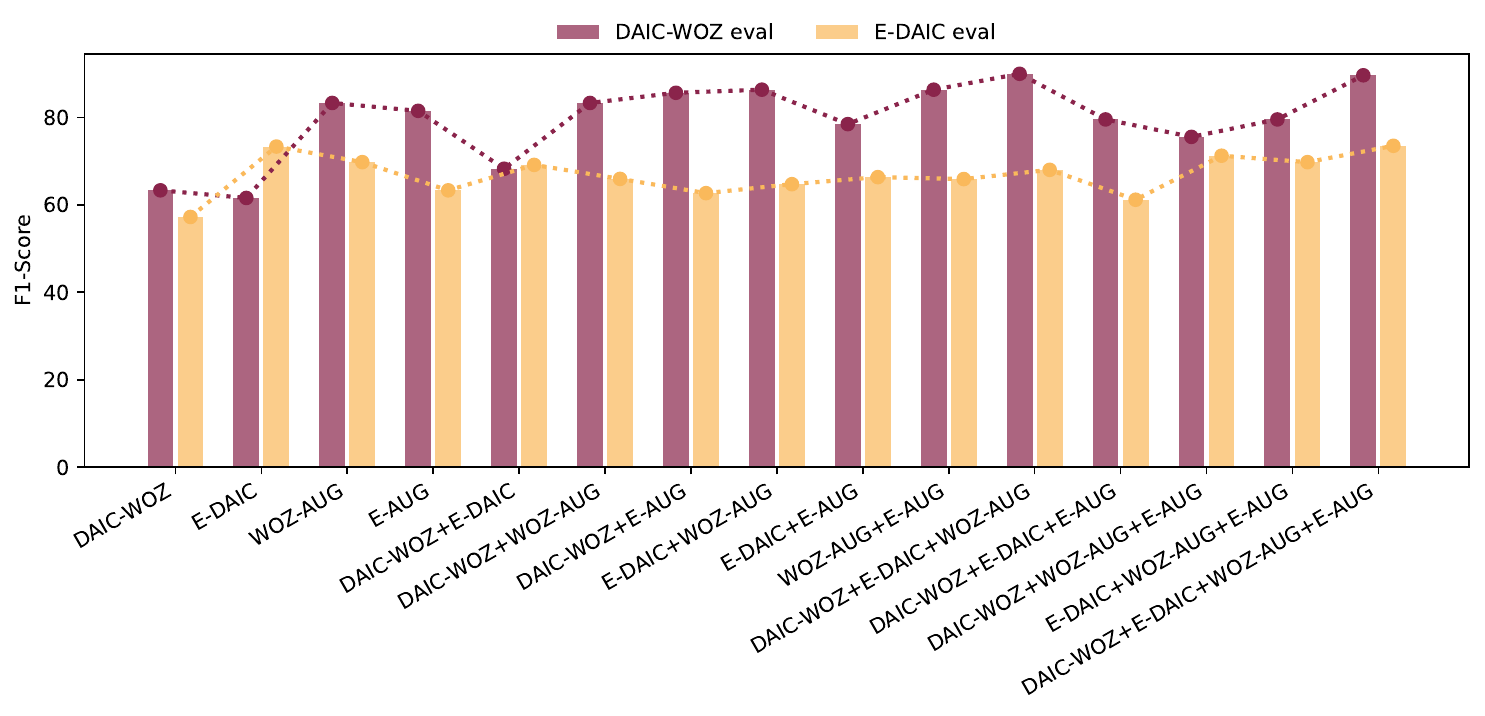}}
\caption{Evaluate performance when training on different dataset configurations}
\label{fig:dataset_combination_ablation}
\end{center}
\end{figure*}

\begin{figure}[ht]
\begin{center}
\centerline{\includegraphics[width=1\linewidth]{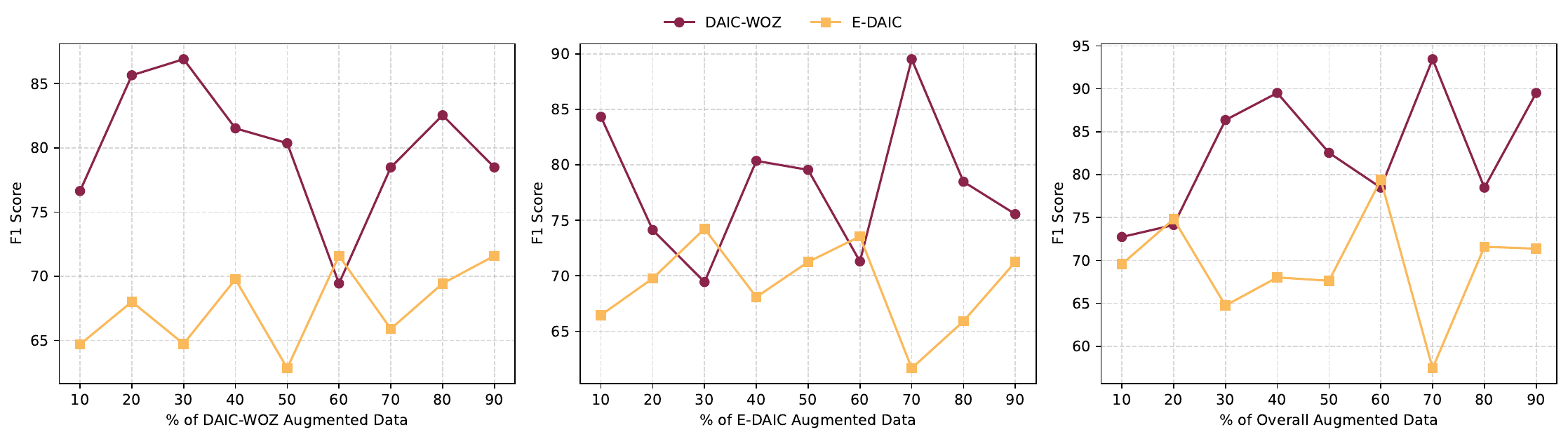}}
\caption{Evaluation on DAIC-WOZ and E-DAIC showing the impact of varying proportions of augmented training data on model F1 performance.}
\label{fig:augmentation_ablation}
\end{center}
\end{figure}





\subsection{Effect of Dataset Composition and Data Augmentation}

We analyze the impact of LLM-based data augmentation from two complementary perspectives: (1) dataset-level training composition and (2) augmentation ratio.

\paragraph{Dataset-Level Training Configurations:}

Figure~\ref{fig:dataset_combination_ablation} shows that training on a single real dataset results in clear generalization gaps. When trained only on DAIC-WOZ, performance reaches approximately 63 F1 on DAIC-WOZ evaluation but drops to about 57 on E-DAIC, indicating limited cross-domain transfer. Similarly, training solely on E-DAIC achieves roughly 73 F1 in-domain but only about 61 on DAIC-WOZ, again revealing dataset-specific overfitting.

Combining the two real datasets (DAIC-WOZ + E-DAIC) improves robustness, yielding approximately 68 F1 on DAIC-WOZ and 69 on E-DAIC, effectively narrowing the cross-domain gap to around 1 point. This confirms that cross-dataset diversity alone improves transferability.

The introduction of augmented data further amplifies these gains. For example, training on WOZ-AUG alone raises DAIC-WOZ performance to approximately 83 F1 (compared to 63 when using only real DAIC-WOZ), while achieving around 70 on E-DAIC. Similarly, mixed configurations such as DAIC-WOZ + WOZ-AUG reach roughly 83--85 F1 on DAIC-WOZ with mid-60s performance on E-DAIC. 

The strongest and most balanced results emerge when real and augmented data from both datasets are combined. The full configuration (DAIC-WOZ + E-DAIC + WOZ-AUG + E-AUG) achieves approximately 89--90 F1 on DAIC-WOZ and about 73--74 on E-DAIC, representing the highest overall performance and one of the smallest cross-domain variances among all settings. Compared to the real-only mixture ($\approx$68/69), this corresponds to gains of more than +20 F1 on DAIC-WOZ and +4-5 F1 on E-DAIC. These results suggest that augmentation acts as a distributional smoothing mechanism, mitigating dataset-specific biases and promoting domain-invariant representations.

\paragraph{Effect of Augmentation Ratio:}

We further examine how varying the proportion of augmented samples influences performance (Figure~\ref{fig:augmentation_ablation}). Across all settings, the relationship between augmentation ratio and performance is clearly non-monotonic.

When progressively increasing DAIC-WOZ augmentation, performance on DAIC-WOZ peaks at around 30\% augmentation ($\approx$87 F1), compared to $\approx$76 at 10\%. Beyond 50-60\%, performance drops substantially ($\approx$69 at 60\%), before partially recovering at higher ratios. A similar pattern is observed when varying E-DAIC augmentation: DAIC-WOZ evaluation peaks at approximately 90 F1 around 70\% E-DAIC augmentation, but lower ratios (20-30\%) yield only mid-70s performance.

When adjusting the overall augmentation proportion across both datasets, the clearest improvements appear in the moderate range (roughly 30-60\%). For instance, at 40\% overall augmentation, DAIC-WOZ evaluation reaches nearly 90 F1, compared to about 73 at 10\%. However, excessive augmentation (e.g., 70\%) leads to degradation, particularly on cross-dataset evaluation, where E-DAIC drops to approximately 57 F1. 

Importantly, performance peaks consistently occur when augmented data constitutes a substantial but not dominant portion of the training set (approximately 30-60\%). When synthetic samples exceed this range, the model appears to over-adapt to LLM-generated distributions, shifting away from authentic conversational characteristics and harming generalization.

Overall, the results demonstrate that LLM-based augmentation significantly enhances robustness and cross-domain transfer when used to complement real data. Moderate augmentation ratios yield improvements of up to +20 F1 compared to single-dataset training, while excessive augmentation introduces distributional noise and performance instability. These findings highlight a critical trade-off: augmentation improves representation diversity and mitigates dataset bias, but only when it remains balanced with authentic data.

\begin{table}[t]
\centering
\small
\setlength{\tabcolsep}{10pt}
\renewcommand{\arraystretch}{1.2}
\begin{tabular}{lcccc}
\toprule
Utterance Span & Hit@1 & Hit@3 & Hit@5 & MRR \\
\midrule
$\pm3$  & \textbf{0.461} & \textbf{0.879} & \textbf{0.990} & \textbf{0.670} \\
$\pm5$  & 0.440 & 0.842 & 0.984 & 0.652 \\
$\pm10$ & 0.395 & 0.815 & 0.964 & 0.620 \\
\bottomrule
\end{tabular}
\caption{Ablation study on utterance span size for causal explanation.
Smaller spans yield more precise causal attribution, while larger spans introduce contextual noise that degrades ranking-based explainability metrics.}
\label{tab:utterance_span_ablation}
\end{table}

\subsection{Explainability over Utterance Spans}

To examine how contextual scope affects causal attribution, we vary the temporal window around each target PEU utterance and report ranking-based explainability metrics in Table~\ref{tab:utterance_span_ablation}. The results show a clear and monotonic degradation as the span increases.

The smallest window ($\pm3$) yields the strongest performance across all metrics, achieving Hit@1 = 0.461, Hit@3 = 0.879, Hit@5 = 0.990, and MRR = 0.670. Expanding the span to $\pm5$ reduces Hit@1 to 0.440 (-0.021) and MRR to 0.652 (-0.018), while Hit@3 and Hit@5 decline to 0.842 and 0.984, respectively. A further expansion to $\pm10$ leads to more pronounced degradation: Hit@1 drops to 0.395 (-0.066 relative to $\pm3$), Hit@3 to 0.815, Hit@5 to 0.964, and MRR to 0.620 (-0.050).

Notably, the decline is sharper for Hit@1 and MRR than for Hit@5. While Hit@5 remains relatively high even at $\pm10$ (0.964 compared to 0.990 at $\pm3$), the larger drop in Hit@1 (from 0.461 to 0.395) indicates that broader spans primarily affect top-ranked precision rather than overall candidate coverage. In other words, the true causal utterance often remains within the extended window, but its ranking position becomes less stable as additional context is introduced.

This pattern suggests that causal signals are predominantly localized within a narrow conversational neighborhood. Enlarging the span introduces competing or less relevant utterances that dilute attribution confidence and interfere with ranking consistency. The progressive decline from $\pm3$ to $\pm10$ therefore reflects diminishing marginal utility of distant context: rather than providing complementary evidence, extended context introduces contextual noise that weakens ranking-based explainability.

Overall, these findings indicate that constraining the causal search space to a focused local window yields more precise and stable causal attribution, reinforcing the importance of localized conversational dynamics in symptom-related reasoning.

\section{Conclusion}
In this work, we introduced \textbf{PsyGAT}, a psychologically grounded graph-based framework for interpretable session-level depression detection from clinical conversations. By formalizing Psychological Expression Units (PEUs) as structured, utterance-level psychological evidence and explicitly modeling their temporal dynamics through directed session graphs, PsyGAT moves beyond surface-level textual cues toward representations aligned with clinically meaningful constructs. To address data scarcity and imbalance in DAIC-WOZ, we incorporated controlled LLM-based augmentation and demonstrated that it improves both in-domain performance and cross-dataset generalization. We further extended PsyGAT with multimodal acoustic features and persona-aware session context, enabling more robust and individualized modeling of depressive patterns across interviews. Beyond prediction, PsyGAT integrates a causal graph-based explainability framework that attributes the emergence of psychological symptoms to preceding conversational context. It achieves strong and stable predictive performance, and rank-based causal explanations that substantially outperform prompt-based LLM baselines. Overall, PsyGAT illustrates how integrating psychological theory, structured graph modeling, and carefully validated LLM components can yield depression detection systems that are accurate, interpretable, robust to distributional shifts, and better aligned with clinical reasoning.

\section*{Acknowledgment}
\noindent
The authors extend their appreciation to the Deanship of Scientific Research and Libraries at Princess Nourah bint Abdulrahman University for funding this work through the Visiting researcher Program VR-2025-003.

\section*{CRediT authorship contribution statement}
\noindent
\textbf{Rishitej Reddy Vyalla:} Conceptualization, Methodology, Software, Validation, Formal analysis, Investigation, Data Curation, Writing - Original Draft, Visualization. \textbf{Kritarth:} Conceptualization, Methodology, Software, Validation, Formal analysis, Investigation, Data Curation, Writing - Original Draft, Visualization. \textbf{Avinash Anand:} Supervision, Project administration, Writing - Review \& Editing. \textbf{Erik Cambria:} Conceptualization, Methodology, Resources, Funding acquisition, Supervision, Project administration, Writing - Review \& Editing. \textbf{Shaoxiong Ji:} Supervision, Writing - Review \& Editing. \textbf{Faten S. Alamri:} Supervision, Writing - Review \& Editing.
\textbf{Zhengkui Wang:} Supervision, Writing - Review \& Editing.

\bibliographystyle{elsarticle-num-names} 
\bibliography{references}







\end{document}